\documentclass[11pt,a4paper]{article}
\usepackage{times,latexsym}
\usepackage{url}
\usepackage[T1]{fontenc}

\usepackage[acceptedWithA]{tacl2021v1}
\usepackage{microtype}
\usepackage{amsmath}
\usepackage{amssymb}
\usepackage{amsfonts}
\usepackage{graphicx}
\usepackage{float}
\usepackage{centernot}
\usepackage{caption}
\usepackage{subcaption}
\usepackage{multirow}
\usepackage{tabularx}
\usepackage{diagbox }
\usepackage{algorithm}
\usepackage{algpseudocode}
\usepackage{soul}

\usepackage{xspace,mfirstuc,tabulary}

\newif\iftaclinstructions
\taclinstructionsfalse 
\iftaclinstructions
\renewcommand{\confidential}{}
\renewcommand{\anonsubtext}{(No author info supplied here, for consistency with
TACL-submission anonymization requirements)}
\newcommand{\instr}
\fi

\iftaclpubformat 

\else

\fi

\newcommand{\system}[1]{\textsc{#1}}
\newcommand{\data}[1]{\textsc{#1}}

\newcommand{\esconv}{\data{ESConv}\xspace}
\newcommand{\msc}{\data{MSC}\xspace}

\newcommand{\ourdataset}{\data{CGDialog}\xspace}

\newcommand{\blenderbot}{\system{BlenderBot}\xspace}
\newcommand{\dialogGPT}{\system{DialoGPT}\xspace}
\newcommand{\roberta}{\system{RoBERTa}\xspace}
\newcommand{\bert}{\system{BERT}\xspace}

\newcommand{\ourmethod}{\system{ConSTrain}\xspace}

\usepackage{amsthm,amsmath,amsfonts,bm,xspace}
\usepackage{color}

\newcommand{\comment}[1]{}

\def\eqref#1{(\ref{#1})}

\def\1{\bm{1}}

\def\rmC{{\mathbf{C}}}

\def\vr{{\bm{r}}}

\def\vt{{\bm{t}}}
\def\vu{{\bm{u}}}

\def\vz{{\bm{z}}}

\DeclareMathAlphabet{\mathsfit}{\encodingdefault}{\sfdefault}{m}{sl}
\SetMathAlphabet{\mathsfit}{bold}{\encodingdefault}{\sfdefault}{bx}{n}

\def\gE{{\mathcal{E}}}

\def\gG{{\mathcal{G}}}

\def\gV{{\mathcal{V}}}

\def\gZ{{\mathcal{Z}}}

\def\sD{{\mathbb{D}}}

\newcommand{\indep}{\perp \!\!\! \perp}
\newcommand{\dep}{\not\!\perp\!\!\!\perp}

\newtheorem{assumption}{Assumption}

\title{Less is More: Mitigate Spurious Correlations for Open-Domain Dialogue Response Generation Models by Causal Discovery}

\author{
  Tao Feng 
  \and
  Lizhen Qu\textsuperscript{*}
  \and
  Gholamreza Haffari 
  \ 
  \\
  Faculty of Information Technology
  \\
  Monash University, Australia
  \\
  \texttt{\{tao.feng, lizhen.qu, gholamreza.haffari\}@monash.edu}
  \\
}

\date{}

\begin{document}
\maketitle
\def\thefootnote{*}\footnotetext{Corresponding author.}\def\thefootnote{\arabic{footnote}}
\begin{abstract}
In this paper, we conduct the \textit{first} study on spurious correlations for open-domain response generation models based on a corpus \ourdataset curated in our work. The current models indeed suffer from spurious correlations and have a tendency of generating irrelevant and generic responses. Inspired by causal discovery algorithms, we propose a novel model-agnostic method for training and inference of response generation model using a conditional independence classifier. The classifier is trained by a constrained self-training method, coined \ourmethod, to overcome data scarcity. The experimental results based on both human and automatic evaluation show that our method significantly outperforms the competitive baselines in terms of relevance, informativeness, and fluency.
\end{abstract}

\section{Introduction}
\begin{table*}[ht]
\centering
\resizebox{\textwidth}{!}{
\begin{tabular}{l|ll|l}
\hline
                          &  Supporter:    & Hello & $\vu_0$                                                            \\ \cline{3-4}
                          & Help seeker: & Hi, how are you? & $\vu_1$                     \\\cline{3-4}
                          & Supporter:   & Doing good.. How are you? & $\vu_2$ \\\cline{3-4}
                          & \multirow{3}{*}{Help seeker:} & \textbf{I'm feeling really anxious these days. I'm finding the COVID online learning experience } \\
                          & & \textbf{to be too much for me at this time.} & $\vu_3$\\
                          & & I want to stop school, but I don't think I can afford to. I need to get done with school.                    &                   \\\cline{3-4}
                          & Supporter:   & I understand your frustration. All of us are challenged due to COVID. & $\vu_4$   \\\cline{3-4}
\multirow{-8}{*}{History} & Help seeker: & School was always hard. Now it's gotten harder. I think a lot of people are stressed. & $\vu_5$      \\ \hline
\begin{tabular}[c]{@{}l@{}}Human\end{tabular} &
  Supporter: &
  \begin{tabular}[c]{@{}l@{}}\textbf{How long are you doing the online school?} \end{tabular} &
   $\vu_6$ \\  \hline
 \blenderbot& Supporter: & You are welcome. I wish you all the best in your future endeavors. Take care. &
  $\vu_7$ \\ \hline
\end{tabular}
}
\caption{An emotion support dialogue annotated with direct causes of human response (in \textbf{bold}). }
\label{tab:spurious_corr_example}
\end{table*}

Open-domain response generation models have achieved impressive empirical success due to the recent advances in large-scale pre-trained transformers~\cite{caldarini2022Chatbotliterature}. However, although those models can generate fluent responses, it is still difficult for them to deeply understand conversation histories, and produce \textit{coherent} and semantically \textit{diverse} responses, especially when the conversation histories are \textit{long}~\cite{sankar2019neuralDialogHistory,qiu2019training}. We conjecture that one of the key reasons is \textit{spuriously correlated} utterances in histories, which do not directly result in responses. Although the vulnerability to \textit{spurious correlations} is a well-known problem in deep learning models~\cite{wang2020CounteringSpuriousCorrelations}, to the best of our knowledge, there is no study on this topic from a causal perspective for response generation models. 

To investigate spurious correlations in dialogues, we are concerned with identifying non-spurious ones, which are the \textit{direct causes} of responses. In this work, a direct cause of a response refers to a text or an utterance in a conversation history that directly results in the response. Table \ref{tab:spurious_corr_example} shows an example dialogue between a help-seeker and a peer-supporter randomly picked from the Emotion Support Conversation corpus (\esconv)~\cite{liu2021emotional}. 
The utterance $\vu_3$ serves as the direct cause of the response $\vu_6$, because it is the only utterance mentioning online learning. Otherwise, if we remove it from the history or significantly alter its meaning, the response $\vu_6$ becomes groundless. In contrast, if we remove an utterance non-causally related to a human response, such as $\vu_1$ or $\vu_5$ related to $\vu_6$, the direct causes still provide sufficient and necessary information to the responses. 


Causal discovery algorithms provide a theoretically grounded way to learn causal relations between random variables from observational data~\cite{nogueira2021causalDiscovery}. Although they can be applied to identify which utterances in conversation histories are direct causes of responses in theory, the research on such methods for natural language processing problems is still in its infancy.

In this work, we conduct the \textit{first} study on spurious correlations for response generation models from a causal perspective. We empirically show that non-cause utterances, including spurious correlated ones, have \textit{significantly more influence} on response generation models than the direct cause utterances human would rely on.

Inspired by causal discovery algorithms, we propose a \textit{model-agnostic} training and inference method for mitigating spurious correlations in long conversations. The method aims to automatically identify key utterances in histories, which serve as direct causes for response generation. Herein we convert the cause identification problem into a problem of conditional independence (CI) tests. The CI tests are realized by building a classifier to infer whether an utterance in the history statistically depends on the response conditioned on its preceding utterance. 
As there is no training data for such a classifier, we start with manually annotating causal relations on a small portion of public open-domain dialogues. To overcome the scarcity of the training data, we propose a \underline{Con}strained \underline{S}elf-\underline{Train}ing method, coined \ourmethod, which is able to identify causal relations with high precision and recall. This classifier is applied to filter out utterances in histories, which are not direct causes of responses, before training response generation models. Furthermore, the classifier serves as a scoring function to select the most relevant response from all generated candidates.     


To sum up, our contributions are as follows: 
\begin{itemize}
\item We conduct the first empirical study on spurious correlations for dialogue response generation models. To investigate this problem in depth, we curate a corpus \ourdataset by annotating causal relations on dialogues.
\item We reduce the direct cause identification problem to a problem of CI tests and propose a constrained self-training method, coined \ourmethod, to train the corresponding classifier.
\item We propose to train response generation models by taking only direct causes as inputs and perform inference using the CI classifier. 
\item The extensive human evaluation results show that the response generation models, such as \blenderbot, using our method outperform the baselines in terms of relevance, informativeness, and fluency. \footnote{ Our dataset, models, and code can be found at \url{https://github.com/WilliamsToTo/CGDIALOG}} 
\end{itemize}

\section{Causal Discovery Background}
\label{sec:background}
Given a set of random variables, causal discovery from observational data is concerned with discovering causal relations between the random variables. A set of causal relations constitutes a causal graph $\gG = \{\gV, \gE\}$, where a node $v \in \gV$ denotes a random variable and a directed edge $v_i \rightarrow v_j \in \gE$ indicates that $v_i$ is a \textit{direct cause} of $v_j$~\cite{neal2020causalitybook}. A change in $v_i$ results in a change in $v_j$, but an intervention in $v_j$ does not necessarily lead to a change in $v_i$. 


Our work is motivated by constraint-based causal discovery approaches~\cite{nogueira2021causalDiscovery}, which iteratively apply independence and CI tests to infer causal structures. Those approaches make the \textit{faithfulness} assumption that independencies in a distribution imply the structure of the corresponding causal graph. The most commonly used algorithm in this family is the PC algorithm~\cite{spirtes2000causation}. It starts with adding undirected edges between two nodes if both of them are dependent by not passing independence tests. Then it remove an  edge between two nodes if they are identified as conditionally independent after running CI tests. The algorithm would continue with larger conditioning sets until the skeleton of the graph is identified. Finally, it orients the edges when possible by using heuristics and identifying the specific structure, $v_i \rightarrow v_k \leftarrow v_j$, referred to as immorality, as illustrated in Fig.\ref{fig:cause_dependence_b}~\cite{neal2020causalitybook}.



In this work, we do not need to recover the complete causal structure between utterances in dialogues. Instead, we only focus on identifying direct causes of responses, namely the parents of the response nodes  in a causal graph. A causal graph satisfies \textit{Causal Markov Condition}, which states that \textit{each variable is independent of all its non-descendants, given its parents in the causal graph}. Hence the value of a response variable is only determined by its parents~\cite{10.5555/3087158.3087202, pearl2009causality}. Under the faithfulness assumption,  if a response variable $v_j$ is dependent on $v_i$ conditioning on arbitrary any other nodes, and we know the influence direction is from $v_i$ to $v_j$, then we conclude that $v_i \rightarrow v_j$. 



\section{Spurious Correlations in Dialogues}
The slogan ``Spurious correlation is no proof of causation'' is well known in statistics~\cite{simon1954spurious}. A correlation between a response and an utterance in a conversational history is spurious, if it does not directly result in the response. 



Spurious correlations are an inherent problem of statistical machine learning (ML) models. \newcite{wang2020CounteringSpuriousCorrelations} point out that ML models relying on core features may well achieve similar training errors on the same training data as those relying on spurious features. However, the models relying on spurious correlations lead to high test errors because spurious correlations are inconsistent across datasets. Overparameterization further exacerbates spurious correlations by memorizing examples containing spurious features~\cite{sagawa2020Overparameterization}. Unfortunately, almost all the SotA open-domain dialogue models are based on large-scale transformers, which are overparameterized with respect to small dialogue training datasets in target domains.


To study the impact of spurious correlations for dialogue models, we leverage two public dialogue corpora (\esconv and \msc) to construct a small evaluation corpus for \underline{C}ausal \underline{G}raphs in \underline{dialog}ues, coined \ourdataset, and evaluate two SotA dialogue models, \blenderbot~\citep{roller2020recipes} and \dialogGPT~\citep{Zhang2020DialoGPTLG}, on that corpus in terms of spurious correlations.

\subsection{Annotation of Causal Graphs}
\label{sec:annotation of causal graphs}
We randomly sampled 80 dialogues from \esconv \citep{liu2021emotional} and \msc \citep{xu2021goldfish} each, then employed four graduate computer science students and four well-trained crowd-workers to annotate direct causes of responses. All annotators were instructed to have a good understanding about what are direct causes of responses and used Amazon Mechanical Turk (AMT) for annotation. We trained them by letting them first annotate on a dry-run dataset, and provided feedback if there was a misunderstanding. After training, annotators were asked to read the provided responses and their conversation histories, then highlight which utterances or clauses serve as direct causes of the responses. We include clause level annotations because sometimes only one clause in a long utterance is the direct cause of a response. For quality check, a human expert having a good grasp of this task reviewed all annotations and corrected mistakes. \ourdataset-\esconv is splitted into a training set, a validation set and a test set, containing 272, 211, 211 context-response pairs respectively, while \ourdataset-\msc contains 300, 250, 250 context-response pairs, respectively.

We measured the inter-annotator agreement between the expert and an annotator at both the utterance level and the clause level.
At the utterance level, we computed Cohen's Kappa and obtained $0.8149$. At the clause level, because marked text boundaries may vary between annotators, we compute the averaged F1 score for all possible pairs of annotators, as detailed in \citet{rajpurkar2016squad,poria2021recognizing}. We obtained a F1 score of $0.8449$, which indicates a high-level of inter-annotator agreement. 

\begin{figure}[h]
    \centering
    \begin{subfigure}[b]{\linewidth}
        \includegraphics[width=0.9\linewidth]{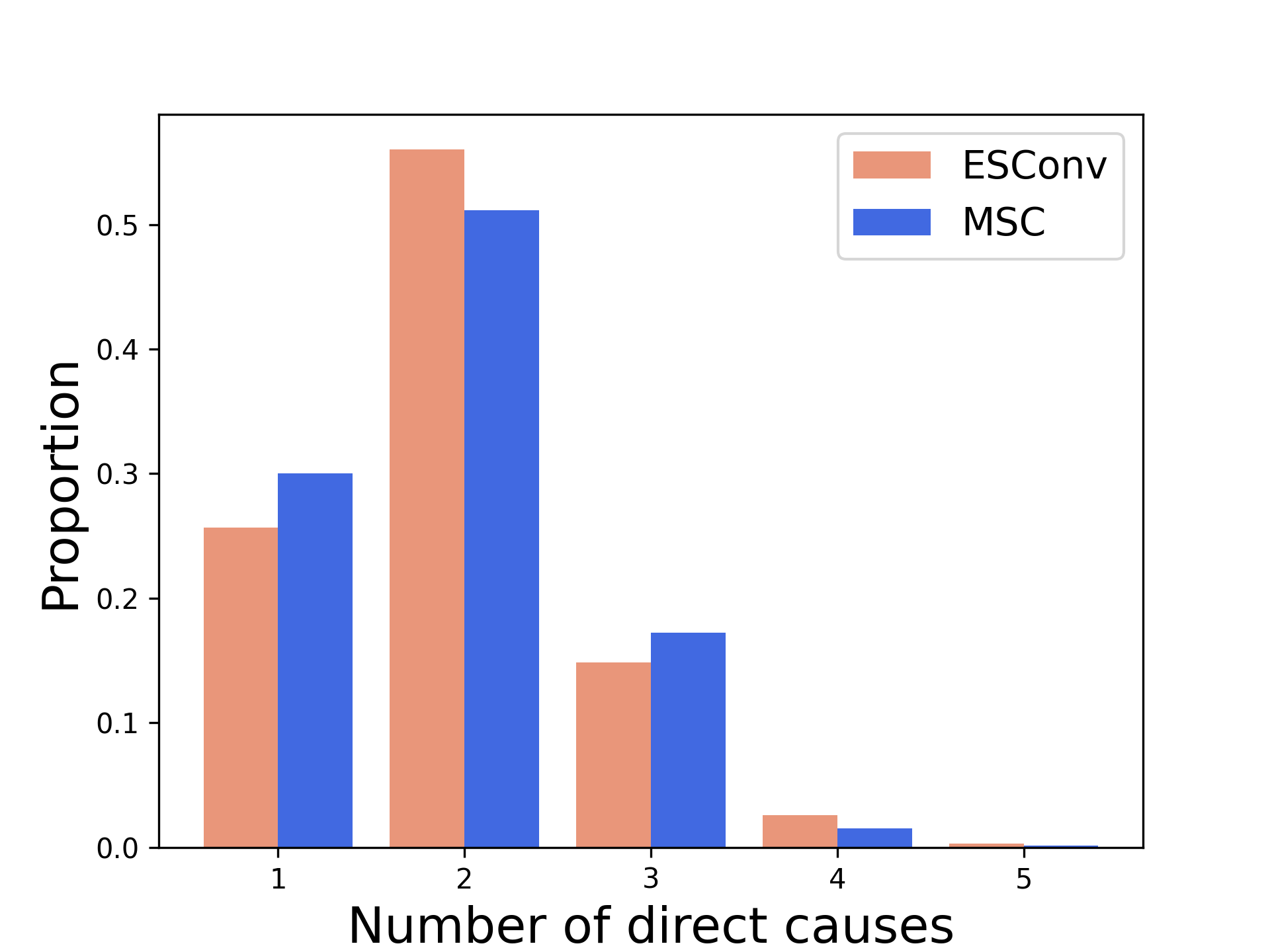}
        \hfill
        \includegraphics[width=0.9\linewidth]{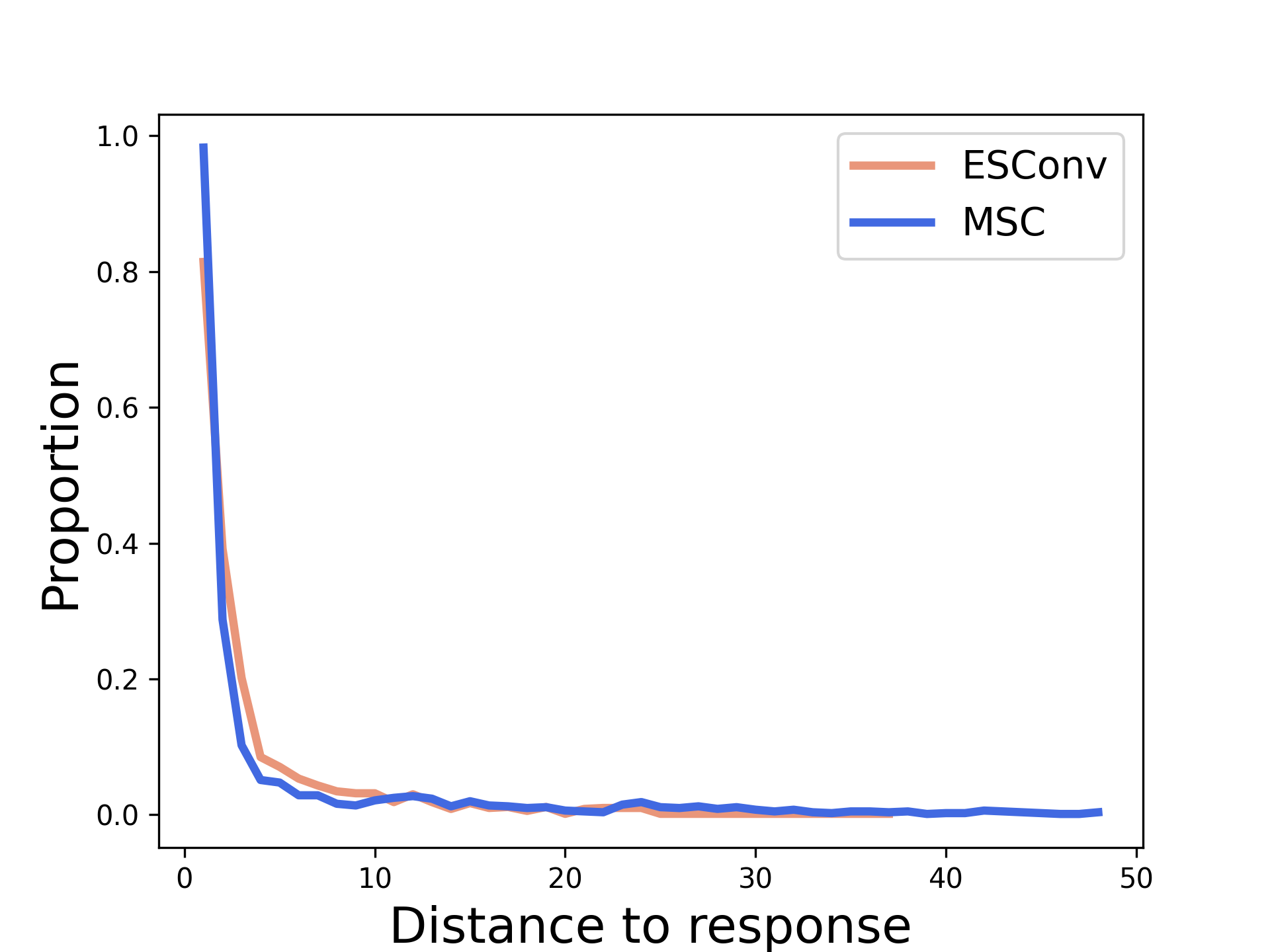}
    \end{subfigure}
    \caption{\textbf{Top:} The ratio between the number of the history-response pairs with a particular number of direct causes and all history-response pairs. \textbf{Down:} Proximity between direct causes and responses, measured by the percentage of such pairs in all history-response pairs. 
    }
    \label{fig: dataset_statistics_figure}
\end{figure}

We show the corpus statistics in Table~\ref{tab:response case statistics} and Figure~\ref{fig: dataset_statistics_figure}. Most of the preceding utterances of responses are annotated as direct causes, which are over $80\%$ and $95\%$ on \esconv and \msc respectively. The proximity of utterances to responses matters: the closer utterances are to the responses, the higher the chance to be direct causes. 

\begin{table}[h]
\centering
\resizebox{\linewidth}{!}{%
\begin{tabular}{ l|lll }
\hline 
\textbf{Number of items}                                         & \textbf{\esconv}             & \textbf{\msc}                & \textbf{Total }             \\  \hline 
Dialogues                                               & 80                 & 80                 & 160                \\ 
History-response pairs                                  & 694                & 800                & 922                \\ 
Utterances                                              & 2301               & 3807               & 6108               \\ 
\begin{tabular}[c]{@{}l@{}}Direct causes utterance\end{tabular}                 & 1347                & 1525                & 2872               \\ 
\begin{tabular}[c]{@{}l@{}}Average token length \\ of direct causes\end{tabular}               & \begin{tabular}[c]{@{}l@{}}24.01\\($\sigma=16.61$)\end{tabular} & \begin{tabular}[c]{@{}l@{}}22.22\\($\sigma=13.79$)\end{tabular} & \begin{tabular}[c]{@{}l@{}}23.05\\($\sigma=15.20$)\end{tabular} \\ 
\begin{tabular}[c]{@{}l@{}}The proportion of direct causes \\ in original utterances \end{tabular} & \begin{tabular}[c]{@{}l@{}}0.86\\($\sigma=0.22$)\end{tabular}   & \begin{tabular}[c]{@{}l@{}}0.72\\($\sigma=0.27$)\end{tabular}   & \begin{tabular}[c]{@{}l@{}}0.79\\($\sigma=0.26$)\end{tabular}   \\ \hline
\end{tabular}
}
\caption{Statistics of the \ourdataset. }
\label{tab:response case statistics}
\end{table}


\subsection{Analysis of Spurious Correlations}
\begin{table*}[h]
\centering
\resizebox{\linewidth}{!}{
\begin{tabular}{llccccccc}
\hline
\multicolumn{1}{l|}{Datasets}                & \multicolumn{1}{l|}{Models}     & \multicolumn{1}{c|}{No Perturbations} & \begin{tabular}[c]{@{}c@{}}Replace non-causes \\ with <pad>\end{tabular} & \begin{tabular}[c]{@{}c@{}} Replace non-causes \\ with <pad> randomly\end{tabular} & \multicolumn{1}{c|}{\begin{tabular}[c]{@{}c@{}}Replace causes \\ with <pad>\end{tabular}} & Drop non-causes & \begin{tabular}[c]{@{}c@{}}Drop non-causes \\ randomly\end{tabular} & Drop causes \\ \hline
\multicolumn{9}{c}{PPL$\downarrow$}                                                                                                                                                                                                                                                 \\ \hline
\multicolumn{1}{l|}{\multirow{2}{*}{ESConv}} & \multicolumn{1}{l|}{Blenderbot} &  \multicolumn{1}{c|}{12.16}            & $25.00^{*}$                   & $12.10$                                    & \multicolumn{1}{c|}{$12.81$}                                                   & $22.65^{*}$        & 13.13          & $12.35$              \\
\multicolumn{1}{l|}{}                        & \multicolumn{1}{l|}{DialoGPT}   &  \multicolumn{1}{c|}{400.15}           & $588.16^{*}$                  & $569.60^{\dagger}$                                   & \multicolumn{1}{c|}{$514.09$}                                                 & $474.42^{*}$        & $469.51^{\dagger}$         & $452.91$             \\
\multicolumn{1}{l|}{\multirow{2}{*}{MSC}}    & \multicolumn{1}{l|}{Blenderbot} &  \multicolumn{1}{c|}{48.29}            & $57.52^{*}$                   & 47.52                                    & \multicolumn{1}{c|}{$49.65$}                                                   & $58.53^{*}$        & 49.69          & $48.82$              \\
\multicolumn{1}{l|}{}                        & \multicolumn{1}{l|}{DialoGPT}   &  \multicolumn{1}{c|}{404.08}           & $875.15^{*}$                 & $703.61^{\dagger}$                                    & \multicolumn{1}{c|}{$613.95$}                                                 & $590.28^{*}$        & $575.12^{\dagger}$         & $480.95$             \\ \hline
\multicolumn{9}{c}{Average BLEU$\uparrow$}                                                                                                                                                                                                                                        \\ \hline
\multicolumn{1}{l|}{\multirow{2}{*}{ESConv}} & \multicolumn{1}{l|}{Blenderbot} &  \multicolumn{1}{c|}{-}                & $0.11^{*}$                      & $0.56^{\dagger}$                                  & \multicolumn{1}{c|}{$0.82$}                                                    & $0.15^{*}$       & $0.48^{\dagger}$            & $0.86$               \\
\multicolumn{1}{l|}{}                        & \multicolumn{1}{l|}{DialoGPT}   & \multicolumn{1}{c|}{-}                & $0.08^{*}$                      & $0.48^{\dagger}$                                  & \multicolumn{1}{c|}{$0.56$}                                                    & $0.11^{*}$       & $0.35^{\dagger}$            & $0.81$               \\
\multicolumn{1}{l|}{\multirow{2}{*}{MSC}}    & \multicolumn{1}{l|}{Blenderbot} & \multicolumn{1}{c|}{-}                & $0.14^{*}$                      & $0.47^{\dagger}$                                  & \multicolumn{1}{c|}{$0.94$}                                                    & $0.09^{*}$       & $0.39^{\dagger}$            & $0.95$               \\
\multicolumn{1}{l|}{}                        & \multicolumn{1}{l|}{DialoGPT}   & \multicolumn{1}{c|}{-}                & $0.28^{*}$                      & $0.49^{\dagger}$                                  & \multicolumn{1}{c|}{$0.81$}                                                    & $0.37^{*}$       & $0.48^{\dagger}$            & $0.82$               \\ \hline
\end{tabular}
}
\caption{Performance comparison with respect to conversation history perturbations. PPL indicates perplexity of human responses. Average BLEU scores are computed as the mean over the four orders of the n-grams. Because responses generated in "No Perturbations" setting are treated as references, average BLEU scores are empty in the "No Perturbations" column. $*$ indicates a significant difference between "Replace (or Drop) causes" and "Replace (or Drop) non-causes", while $\dagger$ represents a significant difference between "Replace (or Drop) causes" and "Replace (or Drop) non-causes randomly". The significant difference is computed by two sample t-test with $p \leqslant 0.05$.}
\label{tab:spurious_correlation_results}
\end{table*}

We conduct experiments to investigate the impact of spurious correlations on two SotA response generation models: \blenderbot and \dialogGPT. Both models are fine-tuned on the training sets of \esconv and \msc by taking full conversation histories as inputs. Inspired by~\cite{sankar2019neuralDialogHistory}, we perturb conversation histories by removing either direct causes or non-causes from histories. We hope that the outputs of a robust model should have little changes if only spuriously correlated utterances are removed. The removal is conducted in two ways: i) replacing each removed token with the pad token <pad>. ii) directly dropping the removed tokens. We apply such perturbations to the test set of \ourdataset and compare their results with the ones without any perturbations.


If a response model captures the same genuine correlations between key utterances in histories and responses as humans, the perplexities of human responses estimated by the model should change only slightly if non-cause utterances are excluded from conversation histories. However, as shown in Table~\ref{tab:spurious_correlation_results}, the increase of perplexities caused by dropping or replacing non-cause utterances is significantly sharper  than that resulted by the removal of cause utterances.

To further investigate the effects of perturbing conversation histories, we apply the same decoding method of both models to the histories after perturbations. We compare the responses generated before and after perturbations in terms of BLEU. Lower BLEU indicates larger changes of generated outputs. As we can see, dropping or replacing direct causes leads to notably smaller changes of outputs than applying the same operations to non-cause utterances. 

To eliminate the concern that the above observations are caused by the number of perturbed utterances, we remove or replace the same number of non-cause utterances as that of direct causes each time. More specifically, as the number of direct causes is always smaller than that of non-causes, we apply the perturbations to $k$ utterances randomly chosen from non-cause utterances if the number of direct causes is $k$, and compute the corresponding perplexities and BLEU. To mitigate the influence of randomness, we repeat each experiment for five times and compute statistical significance based on two-sample t-test~\citep{statisticaltest}. As you can see from Table \ref{tab:spurious_correlation_results}, both generative models are sensitive to the removal of utterances that are weakly associated with human responses. The perturbations on the equal number of non-cause utterances lead to larger changes of the model outputs than those on causes, as indicated by BLEU. For \dialogGPT, the increase of perplexities by perturbing non-causes is still significantly higher than that by perturbing causes. Therefore, both models do not really learn on the utterances that humans use as causes to articulate responses, but rely heavily on non-cause utterances.

\section{Response Generation Based on Causal Discovery}
\label{sec:method}
As shown by our empirical study, spurious correlations are detrimental to the SotA dialogue models. To remedy this, we propose to automatically identify the utterances in conversation histories, which serve as direct causes to responses, and only use them as history representations during both training and inference. Based on the theoretical analysis in Sec. \ref{sec:background}, this identification problem is reduced to running CI tests between responses and utterances in their history. Herein, we propose a constrained self-training procedure to build a classifier for classifier-based CI tests~\cite{c2st2016lopez, sen2017modelpowered, sen2018mimic, bellot2019conditional}. 

Formally, given a conversation history $\rmC_t = \{\vu_0, ..., \vu_{t-1}\}$ at time $t$, a dialogue model aims to produce a word sequence $\vr_{t}$ as the response based on $\rmC_t$. Both $\vu_i$ and $\vr_t$ are regarded as collections of random variable, where each variable in the collection denotes if a single word is present or not. Because the same event can be expressed in various linguistic forms, we assume there is a projection function $g(\vu)$, which maps an utterance to a latent random variable vector $\vz \in \gZ$ denoting the meaning of the corresponding event.  

A causal graph in the semantic space is a directed acyclic graph $\gG = \{\gV, \gE\}$, where a node represents a latent random variable vector $\vz_i$ and an edge is denoted by a causal relation between a pair of nodes. We do not define causal graphs in the word space because i) it is the meanings of utterances that are causally correlated and ii) the same words in different contexts may be involved in different causal relations. Identifying direct causes of responses can thus be regarded as recognizing causal relations between those latent random variables. To simplify notation, we denote the output of $g(\vu_i)$ by $\vz_i$, unless stated otherwise.

\subsection{From Cause Identification to the Conditional Independence Tests}
If a latent semantic vector $\vz_i$ of an utterance is a direct cause of the meaning of a response $\vz_j$, then $\vz_i \dep \vz_j | \mathcal{Z}_{t,-i} $, where $\mathcal{Z}_{t,-i}$ denotes any subset of latent random variables derived from the history $\rmC_t$ excluding $\vz_i$. In other words, $\vz_i$ provides additional useful information for $\vz_j$ given any other utterances in a history. However, it is computationally expensive to consider all possible subsets of a conversation history for running CI tests for a single utterance.

To address the computational challenge, we observe that a response often only depends on the preceding utterance and at most two utterances in total. As evident in Fig.~\ref{fig: dataset_statistics_figure}, $81\%$ of the responses in \ourdataset have one or two direct causes and $90\%$ of the preceding utterances serve as direct causes of the following responses. Therefore, we can sharply reduce the computational overhead by making the following assumptions. 

\begin{assumption}
\label{assume:preceding_cause}
 For each response $\vr_{t}$, $g(\vu_{t-1}) \rightarrow g(\vr_{t})$ always holds.
\end{assumption}

\begin{assumption}
\label{assume:two_causes}
 There are at most two direct causes for the latent random variable vector of a response.
\end{assumption}

\begin{assumption}
\label{assume:orientation}
If there is an edge between $g(\vu_{i})$ and $g(\vu_{j})$ in a causal graph and $i < j$, then $g(\vu_{i}) \rightarrow g(\vu_{j})$.
\end{assumption}

The last assumption articulates the fact that what people said in the past influences what people will say in the future. If the temporal order in a conversation is known, there is no need to apply statistical methods to infer the orientation. 

\begin{figure}[t]
    \centering
    \begin{subfigure}[b]{0.55\linewidth}
        \includegraphics[width=\linewidth]{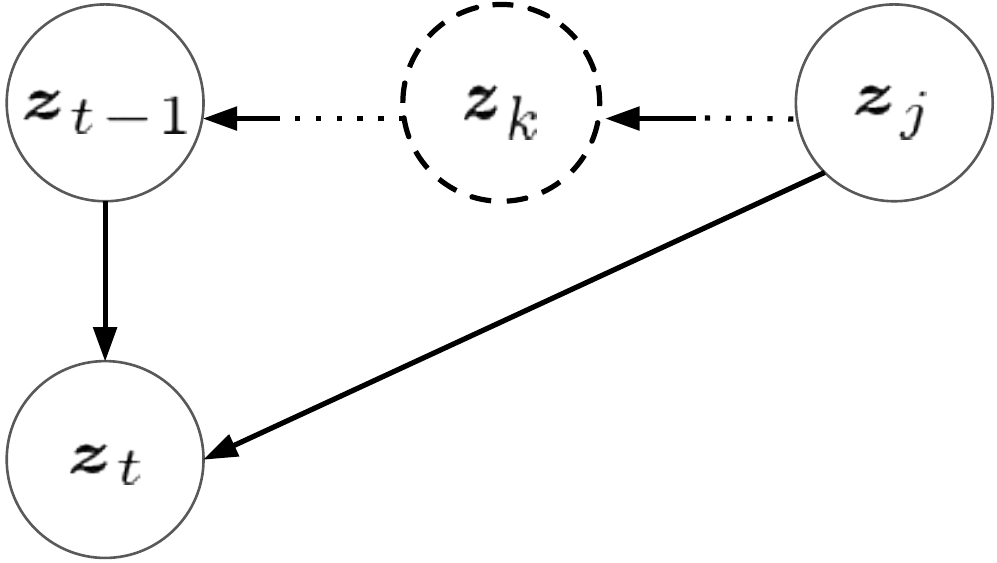}
        \caption{\label{fig:cause_dependence_a}}
    \end{subfigure}
    \hfill
    \begin{subfigure}[b]{0.35\linewidth}
        \includegraphics[width=\linewidth]{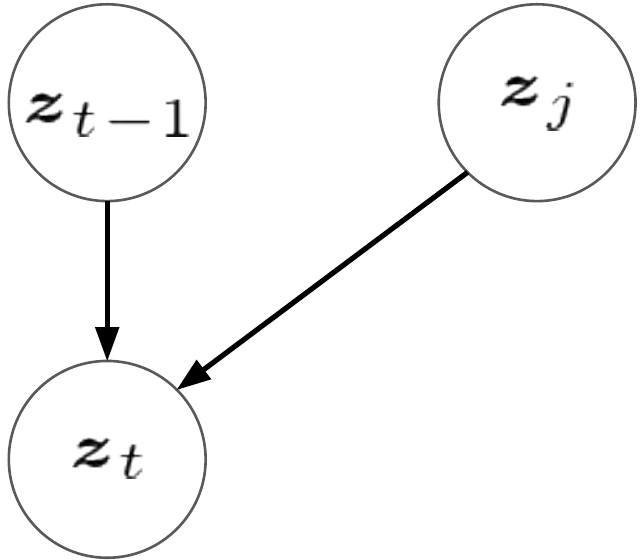}
        \caption{\label{fig:cause_dependence_b}}
    \end{subfigure}
    
    \begin{subfigure}[b]{0.55\linewidth}
        \includegraphics[width=\linewidth]{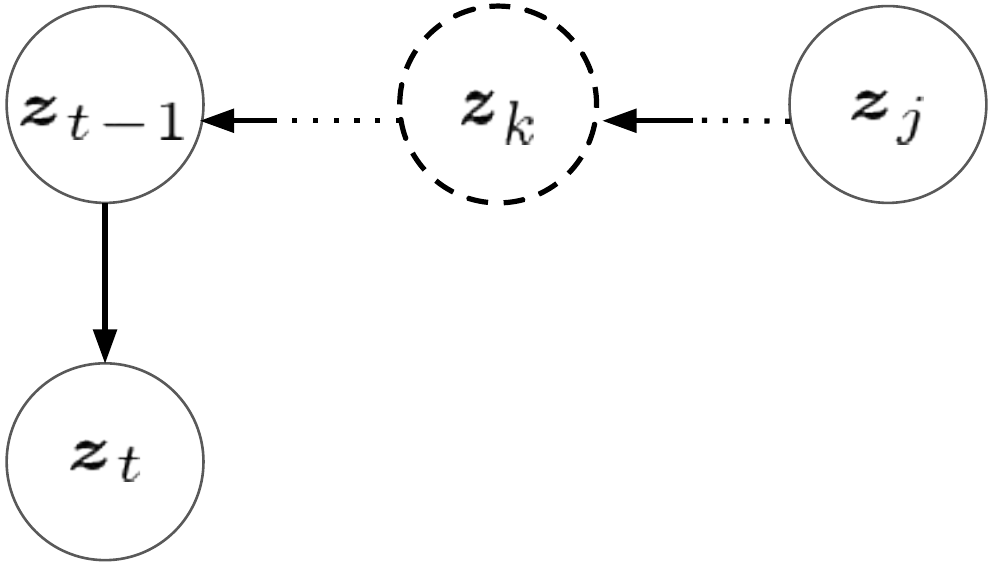}
        \caption{\label{fig:cause_dependence_c}}
    \end{subfigure}
    \hfill
    \begin{subfigure}[b]{0.35\linewidth}
        \includegraphics[width=\linewidth]{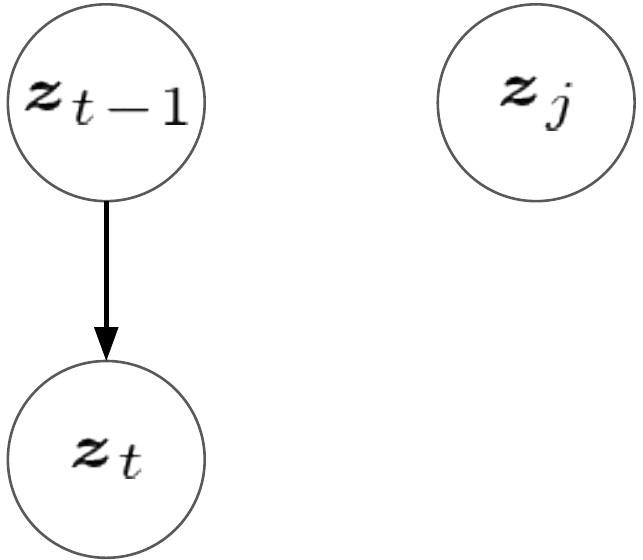}
        \caption{\label{fig:cause_dependence_d}}
    \end{subfigure}
    \caption{In Fig.~\subref{fig:cause_dependence_a}, the response variable has two direct causes that may be connected through $\vz_k$ ($k>j$) or directly connected, while the response variable in Fig.~\subref{fig:cause_dependence_b} has two disconnected cause variables. In Fig.~\subref{fig:cause_dependence_c} and \subref{fig:cause_dependence_d} there is only one direct cause $\vz_{t-1}$ linking to $\vz_t$.}
    \label{fig:real_cause_and_dependence}
\end{figure}

Under the above assumptions, for a given response $\vr_t$, there are only four possible neighborhood structures, as illustrated in Fig.~\ref{fig:real_cause_and_dependence}. We have $\vz_t \dep \vz_j | \vz_{t-1}$ for Fig.\ref{fig:cause_dependence_a} and Fig.\ref{fig:cause_dependence_b}, but $\vz_t$ is conditionally independent of $\vz_j$ in the remaining cases. Herein, we make the faithfulness assumption that CIs imply graph structures. Under our assumptions, it is sufficient to determine if an utterance $\vu_j$ with $j < t$ is a cause of $\vt_t$ by checking whether $\vz_t \dep \vz_j | \vz_{t-1}$. Hence, we only need to run $t-2$ CI tests for a response $\vr_t$. Note that it is \textit{important} to run CI tests instead of dependence tests to find a direct cause of a response. As illustrated in Fig.\ref{fig:cause_dependence_c}, although $\vz_j$ is not a direct cause of $\vz_t$, both of them are still dependent through $\vz_k$ and $\vz_{t-1}$ according to dependence tests. If we run a CI test conditioned on $\vz_{t-1}$, the path through $\vz_k$ is blocked so that the test result reveals $\vz_t \indep \vz_j | \vz_{t-1}$. More details of identifying independence structures in a graphical model can be found in~\citep{neal2020causalitybook, pearl2009causality}.



\subsection{Conditional Independence Tests}
\label{sec: conditional independence test}



To perform CI tests over a set of latent random variables $\vz$ on observational data, we need to i) project utterances to the latent space, and ii) choose a scalable test method which can work with texts. However, the first step is already challenging because the latent random variables are unknown and we even do not know the number of them for an arbitrary dialogue corpus. 

To tackle both challenges, we opt for the classifier-based CI test. As $\vz_t \indep \vz_j | \vz_{t-1}$ implies $p(\vz_t, \vz_j| \vz_{t-1}) = p(\vz_t| \vz_{t-1}) p(\vz_j| \vz_{t-1})$, this family of tests builds a classifier to determine if a sample of data is drawn from $p(\vz_t| \vz_{t-1}) p(\vz_j| \vz_{t-1})$ or $p(\vz_t| \vz_j, \vz_{t-1}) p(\vz_j| \vz_{t-1})$. To train the classifier, we label a tuple $(\vz_t, \vz_{t-1}, \vz_j)$ with $l=1$ if it is drawn from $p(\vz_t| \vz_j, \vz_{t-1}) p(\vz_j| \vz_{t-1})$, otherwise $l=0$. Then the classifier aims to capture the conditional distribution $p(l | \vz_t, \vz_{t-1}, \vz_j)$. 

The recent advances of deep learning show that hidden representations of deep neural networks can well capture meanings of input texts~\cite{yang2020measurement}. Hence, it is straightforward to consider a deep encoder as a function $g(\vu)$ from an utterance $\vu$ to a hidden representation $\vz$. Specifically, we employ a pre-trained \roberta~\cite{liu2019roberta} as the encoder to map a tuple $(\vr_t, \vu_{t-1}, \vu_j)$ to a sequence of hidden representations $(\vz_t, \vz_{t-1}, \vz_j)$, where adjacent utterances are separated by the special token </s>. Taking the representations $(\vz_t, \vz_{t-1}, \vz_j)$ as input, the CI classifier consists of a mean-pooling layer, a linear layer and a sigmoid layer for characterizing $p(l | \vz_t, \vz_{t-1}, \vz_j)$.  


Inspired by~\citet{sun2019unsupervised}, we first train the pre-trained \roberta with the masked language model objective on the publicly available Reddit dataset \citep{baumgartner2020pushshift} to adapt it to dialogues. After training 10 epochs with the learning rate $5\times10^{-5}$, we fine-tune the model with our self-training procedure detailed below. 

\paragraph{Incremental Self-training with Constraints.} It is straightforward to collect a small training dataset $\sD_{L}$ from the training set of \ourdataset by considering $(\vu_j, \vu_{t-1}, \vr_t)$ annotated with $g(\vu_j) \rightarrow g(\vr_t)$ as positive examples and the remaining as negative examples. However, the size of $\sD_{L}$ is small by having only 922 examples in total. 

To address the scarcity of $\sD_{L}$, we adapt the self-training procedure introduced in~\citet{zou2019confidenceSelfTrain} to train the CI classifier. It starts with training an initial classifier $f_0$ on $\sD_{L}$ in a supervised manner. Then we apply this classifier to unlabeled utterance tuples. The tuples predicted with labels 1 are added to the training set as positive examples if they satisfy the \textit{threshold} and \textit{context} constraints :
\begin{enumerate}
    \item[i)] The probability $p(l=1 | \vu_j, \vu_{t-1}, \vr_t )$ exceeds a predefined threshold 0.9;
    \item[ii)] $\vu_j$ is either $\vu_{t-2}$ or $\vu_{t-3}$ with respect to a response $\vr_t$.
\end{enumerate}
For each response $\vr_t$, negative examples are collected by randomly sampling $\vu_j$ from the utterances that are not selected as positive examples. We keep the number of positive examples the same as the number of negative examples in each batch. The extended training set is used to fine-tune the classifier. The process is repeated until the classifier achieves the highest performance on the validation set of \ourdataset. More details can be found in Algorithm \ref{alg: incremental self-training}. Note that, the main difference to the original self-training algorithm is that we add a positive example to the training set only if $\vu_j$ is either $\vu_{t-2}$ or $\vu_{t-3}$. The constraint is proven to be empirically useful in our experiments.

\begin{algorithm}
\caption{Incremental Self-training}
\label{alg: incremental self-training}
\begin{algorithmic}[1]
\Require 
\State Labeled training and validation set: $\sD_{L}^{tr}$, $\sD_{L}^{va}$
\State Unlabeled dataset: $\sD_{U}$
\State Pseudo-labeled data selection constraint: $C$
\State Classifier with pre-trained \roberta: $f_{\theta}$
\Ensure 
\State $i \gets 0$
\State $\sD^{i} \gets \sD_{L}^{tr}$
\State $f_{i} \gets $ fine-tuning $f_{\theta}$ on $\sD^{i}$ 
\While{$f_{i}$ does not have the best performance on $\sD_{L}^{va}$}
\State Apply $f_{i}$ on unlabeled dataset $\sD_{U}$ 
\State Construct pseudo-labeled dataset $\sD^{i}_{PL}$ with constraint $C$
\State $\sD^{i+1} \gets \sD^{i} \cup \sD^{i}_{PL}$
\State $f_{i+1} \gets$ fine-tuning $f_{i}$ on $\sD^{i+1}$ 
\State $i \gets i+1$
\EndWhile
\end{algorithmic}
\end{algorithm}

\subsection{Training and Inference for Generative Response Models}
To overcome spurious correlations, we propose to feed only direct causes of responses to dialogue models during training and inference, where direct causes are selected by the CI classifier. This approach is model-agnostic because it only ``cleans'' the inputs of a response model regardless which neural architecture is used.

The training set of mainstream open-domain dialogue models consists of conversation history and response pairs $\{\rmC_t, \vr_t\}_{t=1}^n$. Before training, we preprocess the training set by keeping only direct causes in each conversation history. As $\vu_{t-1}$ is always one of the direct causes according to Assumption \ref{assume:preceding_cause}, we find another cause by using the CI classifier. In particular, for each conversation history $\rmC_t$, we perform max inference on all tuples $(\vu_{j}, \vu_{t-1}, \vr_{t})$ using the classifier, where $j\in [0, t-2]$. We select the $\vu_j$ that has the highest probability $p(l=1\mid \vu_{j}, \vu_{t-1}, \vr_{t})$ as another direct cause. Dialogue models are subsequently trained on the preprocessed training set.

The input selection for inference is conducted in a similar manner. In particular, we feed each possible $(\vu_j, \vu_{t-1})$ with $j\in [0, t-2]$ to the trained dialogue model to generate a response by beam search. Then we apply the CI classifier to identify the tuple $(\vu_j, \vu_{t-1}, \vr_{t})$ with the highest $p(l=1\mid \vu_{j}, \vu_{t-1}, \vr_{t})$. To allow selecting responses based on $p(\vr_{t} | \vu_j, \vu_{t-1})$ or $p(\vr_{t}|\vu_{t-1})$, we choose the response conditioned on $(\vu_j, \vu_{t-1})$ if the highest $p(l=1\mid \vu_{j}, \vu_{t-1}, \vr_{t})$ exceeds the threshold $0.5$, tuned on a validation set, otherwise we take the response conditioned on $\vu_{t-1}$. 



\section{Experiments}
\label{sec:exp}
\subsection{Datasets}
\label{sec: datasets}
We experiment on the following two open-domain dialogue corpora that have long conversation histories. The longer a conversation history is, the more likely utterances in the history are spuriously correlated with responses. In contrast, most open-domain dialogue corpora contain short conversations, in which there are dramatically less spuriously correlated utterances. For example, DailyDialog \cite{li-etal-2017-dailydialog}, WizardOfWikipedia \cite{dinan2019wizard} and EmpatheticDialogues \cite{rashkin2019empathetic} have 7.9 utterances, 9 utterances, and 4.31 utterances per conversation, respectively. 
\paragraph{Emotion Support Conversation (\esconv).} \esconv \citep{liu2021emotional} contains conversations between mental health help seekers and supporters, with $29.8$ utterances per dialogue on average. In each dialogue, help seekers talk about their problems, such as unemployment, losing family member or infecting with COVID. Dialogue response models play the role of supporters to provide supportive responses to help seekers. Each utterance from supporters is annotated with a strategy such as providing suggestions, paraphrasing or question, which are not considered in our models. 
It is splitted into training, validation and test sets with the ratios of $80\%$, $10\%$ and $10\%$ respectively.

\paragraph{Multi-Session Chat (\msc).} \msc \citep{xu2021goldfish} contains human-human chit-chats over five sessions, each of which contains up to 14 utterances. 
The average number of utterances per dialogue is $53.3$. In each session, two interlocutors conduct a conversation based on given personas. Each persona describes personal information with multiple sentences. 
We experiment on its official splits of training, validation, and test sets. 

\subsection{Baseline Models}
We compare our method \ourmethod and its variations, based on \blenderbot, with the following generative models: 
\paragraph{\blenderbot.} 
This transformer-based encoder-decoder model 
achieves superior performance over the prior models in terms of engagingness and humanness    \citep{roller2020recipes}. We fine-tune the pre-trained model with varying settings of conversational histories. As such, a conversational history contains either:
1) only the preceding utterance $\vu_{t-1}$, 
2) the preceding two utterances $(\vu_{t-2}, \vu_{t-1})$ when available, 
3) the preceding three utterances $(\vu_{t-3}, \vu_{t-2}, \vu_{t-1})$ when available, 
4) the complete conversational history $(\vu_0, ..., \vu_{t-1})$, or
5) the preceding utterance $\vu_{t-1}$ and a randomly selected utterance $\vu_j$ between $0$ and $t-2$. 
All hyperparameters remain the same in different settings. 

\paragraph{DialoFlow.}
\citet{li2021dialoflow} proposes a dialogue system that models dynamic information flow across utterances. The model generates a response based on a distributed representation predicted based on past information flow.

\paragraph{Retrieval-guided Model.}
We implement the retrieval-guided response generation model proposed in \citep{zhong-etal-2022-less} without using user ids, because they are not available in both corpora. Herein, we first map the tokens in the preceding utterance $\vu_{t-1}$ and the tokens in the previous history $\{\vu_0,..., \vu_{t-2}\}$ into a set of \bert embeddings respectively. Then we compute a similarity matrix between the two sets of embeddings in terms of dot product. As there is a similarity vector for each token in the previous history, we score each of them by using the highest similarity score in the corresponding vector. We pick the top 30 scored ones as the final set of retrieved tokens. 
The input to their response generation model is the concatenation of $\vu_{t-1}$ and the corresponding retrieved tokens. 

\paragraph{\esconv Baseline.} 
\citet{liu2021emotional} provide two response models on \esconv. The first one directly fine-tunes the \blenderbot model on \esconv without using annotations of negotiation strategies. Another one fine-tunes \blenderbot by taking as input both negotiation strategies and conversation histories. Both models consider the preceding five utterances as conversation history. 

\paragraph{TransferTransfo.}
As \msc can be viewed as an extension of PersonaChat dataset \citep{zhang2018personalizing}, we consider TransferTransfo \citep{wolf2019transfertransfo}, which reports the SotA performance on PersonaChat. We fine-tune this model on the training set of \msc for a fair comparison.

\paragraph{Retriever-generator.}
\citet{xu2021goldfish} propose a model consisting of a retriever and a generator. The retriever selects relevant utterances from a history, while the generator produces responses conditioned on the utterances selected by the retriever.

Amongst the above models, \blenderbot, DialoFlow, and retrieval-guided model are evaluated on both corpora. TransferTransfo is evaluated only on \msc because the same model shows inferior performance than the one proposed in \citet{liu2021emotional} on \esconv. Furthermore, the baseline \citep{liu2021emotional} is only evaluated on \esconv because it requires annotations of strategies. 

\subsection{Implementation Details}
All the models are implemented with PyTorch \citep{paszke2019pytorch} and the Transformers library \citep{wolf-etal-2020-transformers}. We use the same \blenderbot model\footnote{\url{https://huggingface.co/facebook/blenderbot-400M-distill}} in all relevant experiments. All models are trained with Adam \citep{kingma2017adam} optimizer with hyperparameters tuned on the validation sets. As a result, we run Adam with $\beta_{1}=0.9$ and $\beta_{2}=0.999$. The learning rate is $2\times 10^{-5}$ for CI classifier and $5\times 10^{-5}$ for the response model. We use a linear learning rate scheduler that dynamically decreases learning rate after a warm-up period. CI classifiers were trained for $10$ epochs with the batch size $16$ on one NVIDIA RTX 16G V100 GPU; the response models were trained with $5$ epochs and a batch size of $8$. The beam search width is set to $5$ during decoding. 

\subsection{Metrics}
\label{sec:metrics}
\paragraph{Human Evaluation}
 In practice, we had the same observations as in \cite{belz-kow-2010-comparing, callison-burch-etal-2007-meta, Kiritchenko2017BestWorstSM} that asking crowd-workers to directly score responses on a scale usually receives low-quality evaluation. Thus, following the evaluation design in \citep{novikova2018rankme, bojar-etal-2016-findings, zheng-etal-2021-comae, zhou2018emotional, liu2021emotional}, we opt for pairwise comparison between responses from different sources. In each comparison experiment, we compared our model with a baseline or human responses on a set of 100 conversations randomly sampled from our test set. Given a conversation history, we presented crowd-workers with a pair of responses, one of which is generated by our model and the other is either from humans or a baseline. Five well-trained crowd-workers from Amazon Mechanical Turk (AMT) are asked to choose the better one in terms of four metrics:  \textbf{Empathy} (Which response shows better understanding of the partner's feelings?), \textbf{Fluency} (Which response has better fluency and readability?), \textbf{Relevance} (Which response is more relevant and coherent to the context?) and \textbf{Informativeness} (Which response provides more information when both are relevant?). For quality control, we selected only crowd-workers who have an approval rating greater than $90\%$ and a minimum of $10,000$ approved tasks. Inter-rater agreement using Krippendorff’s $\alpha$ was 0.41. In addition, we presented both good and bad example responses for each metric to educate crowd-workers. 
 
 The results of all comparison experiments are summarized by using ranking-based Best-Worst Scaling, a method shown to be more reliable than rating-based Likert scaling in prior studies~\cite{Kiritchenko2017BestWorstSM, puduppully-lapata-2021-data, Steen2021HowTE, Tang2022InvestigatingCP, louviere2015best}. For each pair of models in comparison, the score of a model is calculated as the number of times rated best minus the number of times rated worst \citep{amplayo-lapata-2021-informative, puduppully-lapata-2021-data}. Thus, for such a pair of models, their scores have the same absolute value but opposite signs. For example, in a comparison experiment between System A and System B, the score of System A is 13, then that of System B is -13. Thus, we only need to know the score of one system, then obtain the score of the other system automatically. To summarize those results, we put the scores of baselines and human responses in one table, which are compared with our model. As our model is always used as a reference, we set the scores of our model to be zero in that table. Therefore, a negative score in the table means the corresponding system performs worse than our model, while a positive score indicates a better performance of the corresponding system.

\paragraph{Automatic Evaluation}
 Although automatic metrics are still not reliable for response evaluation~\citep{liu2017evaluate}, to facilitate comparisons with prior works, we consider the four automatic metrics for evaluating the quality of responses: BLEU \cite{Papineni2002BleuAM}, BERTScore \cite{Zhang2020BERTScoreET}, MAUVE \cite{pillutla2021mauve}, METEOR \cite{banarjee2005}. In addition, we evaluate the diversity of model outputs in terms of Distinct-1/2 \cite{Li2016ADO}.
\subsection{Experimental Results}

\begin{table*}[h]
\centering
\resizebox{0.8\linewidth}{!}{
\begin{tabular}{lcccc}
\hline
\multicolumn{1}{l|}{\textbf{Models}} & \textbf{Empathy}$\uparrow$ & \textbf{Fluency}$\uparrow$ & \textbf{Informativeness}$\uparrow$ & \textbf{Relevance}$\uparrow$ \\ \hline
\multicolumn{5}{c}{\esconv}        \\ \hline
\multicolumn{1}{l|}{\blenderbot - $P(\vr_{t} | \vu_{t-1})$} & -22* & -48* & -15* & -4 \\
\multicolumn{1}{l|}{\blenderbot - $P(\vr_{t} | \vu_{t-2:t-1})$} & -83* & -46* & -12 & -26* \\
\multicolumn{1}{l|}{\blenderbot - $P(\vr_{t} | \vu_{t-3:t-1})$} & -28* & -39* & -31* & -31* \\
\multicolumn{1}{l|}{\blenderbot - $P(\vr_{t} | \vu_{0:t-1})$} & -54* & -36* & -16* & -38* \\
\multicolumn{1}{l|}{\blenderbot - $P(\vr_{t} | \vu_{j}, \vu_{t-1})$} & -69* & -61* & -25* & -51* \\
\multicolumn{1}{l|}{DialoFlow} & -38* & -54* & -6 & -28* \\
\multicolumn{1}{l|}{\citep{liu2021emotional} w/o strategy} & -64* & -45* & -6 & -9 \\
\multicolumn{1}{l|}{\citep{liu2021emotional} with strategy} & -52* & -36* & -13* & -19* \\
\multicolumn{1}{l|}{Retrieval-guided} & -3 & -14* & -12* & -18* \\ 
\multicolumn{1}{l|}{\ourmethod (Ours)}               & 0    & \textbf{0}    & \textbf{0}            & 0      \\ 
\multicolumn{1}{l|}{Human}               & \textbf{12}    & -30*    & -16*            & \textbf{3}      \\
\hline
\multicolumn{5}{c}{\msc}           \\ \hline
\multicolumn{1}{l|}{\blenderbot - $P(\vr_{t} | \vu_{t-1})$} & -    & -31* & -25* & -7 \\
\multicolumn{1}{l|}{\blenderbot - $P(\vr_{t} | \vu_{t-2:t-1})$} & -    & -54* & -24* & -35* \\
\multicolumn{1}{l|}{\blenderbot - $P(\vr_{t} | \vu_{t-3:t-1})$} & -    & -12 & -8 & -4 \\
\multicolumn{1}{l|}{\blenderbot - $P(\vr_{t} | \vu_{0:t-1})$} & -    & -80* & -30* & -80* \\
\multicolumn{1}{l|}{\blenderbot - $P(\vr_{t} | \vu_{j}, \vu_{t-1})$} & -    & -82* & -71* & -66* \\
\multicolumn{1}{l|}{DialoFlow} & - & -54* & -35* & -51* \\
\multicolumn{1}{l|}{TransferTransfo} & -    & -49* & -44* & -48*  \\
\multicolumn{1}{l|}{Retriever-generator} & -    & -64* & -10 & -14 \\
\multicolumn{1}{l|}{Retrieval-guided} & - & -12 & -29* & -32* \\
\multicolumn{1}{l|}{\ourmethod (Ours)}               & -                & 0    & 0            & 0      \\ 
\multicolumn{1}{l|}{Human} & -    & \textbf{3} & \textbf{19}* & \textbf{19}*  \\
\hline
\end{tabular}
}
\caption{Results of human evaluation using best-worst scaling (higher is better). The results in \textbf{Bold} are better than all the competitors. Systems significantly different from our method are marked with an asterisk * (using a one-way ANOVA with post hoc Tukey HSD tests; $p \leqslant  0.05$).}
\label{tab:human evaluation result}
\end{table*}

\begin{table*}[h]
\centering
\resizebox{\textwidth}{!}{
\begin{tabular}{lcccccc}
\hline
\multicolumn{1}{l|}{\textbf{Models}}    &    \textbf{BLEU}$\uparrow$  &  \textbf{BERTScore}$\uparrow$    & \textbf{MAUVE}$\uparrow$  &\textbf{METEOR}$\uparrow$    & \textbf{D-1}$\uparrow$ & \textbf{D-2}$\uparrow$   \\ \hline
\multicolumn{7}{c}{\esconv}        \\ \hline
\multicolumn{1}{l|}{\blenderbot - $P(\vr_{t} | \vu_{t-1})$}                  &    0.09 & 0.19  & 0.24   & 0.12 & 0.26      & 0.72 \\  
\multicolumn{1}{l|}{\blenderbot - $P(\vr_{t} | \vu_{t-2:t-1})$}             &   0.09 & 0.19    & 0.32    & 0.12 & 0.27      & 0.73 \\ 
\multicolumn{1}{l|}{\blenderbot - $P(\vr_{t} | \vu_{t-3:t-1})$}             &    0.08 & 0.18    & 0.24    & 0.13 & 0.27      & 0.73 \\ 
\multicolumn{1}{l|}{\blenderbot - $P(\vr_{t} | \vu_{0:t-1})$}                & 0.08 &  0.15    & 0.09    & 0.11 & 0.27      & 0.73 \\ 
\multicolumn{1}{l|}{\blenderbot - $P(\vr_{t} | \vu_{j}, \vu_{t-1})$}          & 0.07 & 0.14   & 0.29   & 0.11 & 0.24      & 0.70 \\ 
\multicolumn{1}{l|}{DialoFlow}                                          & 0.05 & 0.14     & 0.19    & 0.07 & 0.23      & 0.72 \\ 
\multicolumn{1}{l|}{\citep{liu2021emotional} w/o strategy}               & 0.09 & 0.18      & 0.31    & 0.12 & 0.24      & 0.70 \\ 
\multicolumn{1}{l|}{\citep{liu2021emotional} with strategy}              & 0.07 & 0.18      & 0.21    & 0.13 & 0.27      & 0.73 \\ 
\multicolumn{1}{l|}{Retrieval-guided}                               & 0.07 & 0.17     & 0.27    & 0.12 & 0.26      & 0.72 \\ 
\multicolumn{1}{l|}{\ourmethod (Ours)}           & 0.08 & 0.18    & 0.33    & 0.13 & 0.26      & 0.73 \\ \hline 
\multicolumn{7}{c}{\msc}           \\ \hline
\multicolumn{1}{l|}{\blenderbot - $P(\vr_{t} | \vu_{t-1})$}              &    0.09 & 0.20    & 0.28    & 0.11 & 0.28      & 0.74 \\ 
\multicolumn{1}{l|}{\blenderbot - $P(\vr_{t} | \vu_{t-2:t-1})$}         &    0.09 & 0.20  & 0.30    & 0.10 & 0.29      & 0.76 \\ 
\multicolumn{1}{l|}{\blenderbot - $P(\vr_{t} | \vu_{t-3:t-1})$}         &    0.08 & 0.18   & 0.23    & 0.11 & 0.29      & 0.76 \\ 
\multicolumn{1}{l|}{\blenderbot - $P(\vr_{t} | \vu_{0:t-1})$}           & 0.06 & 0.13     & 0.02   & 0.08 & 0.26      & 0.75 \\ 
\multicolumn{1}{l|}{\blenderbot - $P(\vr_{t} | \vu_{j}, \vu_{t-1})$}       & 0.07 & 0.16    & 0.07    & 0.09 & 0.27      & 0.74 \\ 
\multicolumn{1}{l|}{DialoFlow}                                           & 0.05 & 0.14     & 0.16    & 0.08 & 0.33      & 0.74 \\ 
\multicolumn{1}{l|}{TransferTransfo}                                 & 0.07 & 0.13    & 0.10     & 0.05 & 0.50      & 0.89 \\ 
\multicolumn{1}{l|}{Retriever-generator}                                & 0.09 & 0.20    & 0.25    & 0.10 & 0.29      & 0.75 \\ 
\multicolumn{1}{l|}{Retrieval-guided}                                  & 0.08 & 0.18      & 0.20    & 0.11 & 0.26      & 0.74 \\ 
\multicolumn{1}{l|}{\ourmethod (Ours)}        & 0.09 & 0.20      & 0.31    & 0.13 & 0.29      & 0.76 \\ \hline
\end{tabular}
}
\captionsetup{width=\textwidth}
\caption{Automatic evaluation results contain BLEU, BERTScore (F1), MAUVE, METEOR, and Distinct (D1 and D2). Distinct score is calculated on 1-gram and 2-gram on corpus level. }
\label{tab:automatic evaluation result}
\end{table*}

\begin{table}[h]
\centering
\resizebox{\linewidth}{!}{
\begin{tabular}{lcccc}
\hline
\multicolumn{1}{l|}{\textbf{Models}} & \textbf{Empa}$\uparrow$ & \textbf{Fluen}$\uparrow$ & \textbf{Info}$\uparrow$ & \textbf{Rele}$\uparrow$ \\ \hline
\multicolumn{5}{c}{\esconv}        \\ \hline
\multicolumn{1}{l|}{$P(\vr_{t} | \vu_{t-1})$} & -3 & -11 & -14* & -21* \\
\multicolumn{1}{l|}{$P(\vr_{t} | \vu_{t-2:t-1})$} & -12 & -17* & -25* & -28* \\
\multicolumn{1}{l|}{$P(\vr_{t} | \vu_{t-3:t-1})$} & -11 & -5 & -25* & -18* \\
\multicolumn{1}{l|}{$P(\vr_{t} | \vu_{0:t-1})$} & -26* & -32* & -22* & -20* \\
\multicolumn{1}{l|}{\ourmethod}               & \textbf{0}    & \textbf{0}    & \textbf{0}            & \textbf{0}      \\ 
\hline
\multicolumn{5}{c}{\msc}           \\ \hline
\multicolumn{1}{l|}{$P(\vr_{t} | \vu_{t-1})$} & -    & -9 & -7 & -11 \\
\multicolumn{1}{l|}{$P(\vr_{t} | \vu_{t-2:t-1})$} & -    & -5 & -10 & -15* \\
\multicolumn{1}{l|}{$P(\vr_{t} | \vu_{t-3:t-1})$} & -    & -16* & -28* & -18* \\
\multicolumn{1}{l|}{$P(\vr_{t} | \vu_{0:t-1})$} & -    & -13* & -23* & -17* \\
\multicolumn{1}{l|}{\ourmethod}               & -                & \textbf{0}    & \textbf{0}            & \textbf{0}       \\ 
\hline
\end{tabular}
}
\caption{Model-agnostic experiment results where all models use \dialogGPT as backbone. * indicates significant difference with \ourmethod - $\vu_{MaxCI, t-1}$.
}
\label{tab:DialoGPT human evaluation result}
\end{table}

\paragraph{Response Generation.}
%
We compare \blenderbot using our method (\ourmethod) with multiple strongest baselines for response generation. Table~\ref{tab:human evaluation result} summarizes the human evaluation results based on the Best-Worst Scaling. Our response model outperforms all baselines in terms of all the metrics on both \esconv and \msc, as indicated by their negative scores. Most of the results are statistically significant. The automatic evaluation results with MAUVE in Table \ref{tab:automatic evaluation result}, one of the best automatic metrics for NLG tasks, also demonstrates the strengths of our method over the baselines. This meets our expectation that responses generated based on direct causes perform better than responses generated on histories including spuriously correlated utterances.

Surprisingly, the \blenderbot using our method outperforms human responses on \esconv in terms of fluency and informativeness. A close look at the results reveals that i) some of the responses generated by our model are longer than the corresponding human responses because they cover more specific details in contexts, and ii) a significant amount of responses in \esconv contain grammatical errors while the model generated ones rarely make grammatical errors. Unfortunately, our model does not reach human-level performance on \msc in terms of informativeness and relevance, in which the majority of the multi-session conversations span more than 40 turns. 

The two model variations in \citet{liu2021emotional} are the reported strongest baselines on \esconv, while the retriever-generator model is the strongest one on \msc in literature. Both the retriever-generator and the retrieval-guided model apply retrieval techniques to identify the most relevant texts in context. The retrieval-guided model starts with employing the tokens in the preceding utterance $\vu_{t-1}$ as queries to retrieve the most relevant tokens in the context $\{\vu_0, ..., \vu_{t-2}\}$, followed by concatenating them with the ones in $\vu_{t-1}$ as model inputs. In contrast, retriever-generator identifies relevant utterances in histories. Despite that, all of them still fall short of our method according to human and automatic evaluations. Those results indicate that retrieval techniques are still limited for identifying key utterances from conversation histories. 

We compare different ways of selecting utterances from conversation histories as the inputs of the same neural architecture. Table~\ref{tab:human evaluation result} and Table \ref{tab:automatic evaluation result} include the corresponding results of \blenderbot on both corpora. Taking the full conversation histories as input, which is widely used in practice, turns out to be a poor choice on both corpora. The responses generated in this setting are often too general, such as ``I'm sorry to hear that.'', without touching specific details in contexts. As a comparison, using the preceding utterances is evident as a good heuristic on \esconv, while the best heuristic on \msc is to use the preceding three utterances. The worse case is $P(\vr_{t} | \vu_{j}, \vu_{t-1})$, which randomly selects an utterance between the first utterance and $\vu_{t-2}$ to combine with $\vu_{t-1}$. The corresponding ratio of spurious correlations is one of the highest among all settings. Those results again demonstrate the harm of spuriously correlated utterances for generative models. 


\begin{table*}[h]
\centering
\resizebox{\linewidth}{!}{
\begin{tabular}{l|cccc|cccc}
\hline
                                                            & \multicolumn{4}{c|}{\esconv}                                                                           & \multicolumn{4}{c}{\msc}                                                                               \\ \hline
\textbf{Models}                                             & \textbf{Empa}$\uparrow$ & \textbf{Fluen}$\uparrow$ & \textbf{Info}$\uparrow$ & \textbf{Rele}$\uparrow$ & \textbf{Empa}$\uparrow$ & \textbf{Fluen}$\uparrow$ & \textbf{Info}$\uparrow$ & \textbf{Rele}$\uparrow$ \\ \hline
\ourmethod (Ours)                                           & 0                       & 0                        & 0                       & 0                       & -                       & 0                        & 0                       & 0                       \\
\ourmethod - $\vu_{t-2, t-1}$                               & -21*                    & 2                        & -6                      & -7                      & -                       & -2                       & 3                       & -4                      \\
\ourmethod - $\vu_{MaxDep, t-1}$                            & -9                      & -13*                     & -14*                    & -10                     & -                       & -17*                     & 4                       & -18*                    \\
\ourmethod - $\vu_{Random, t-1}$                            & -22*                    & -18*                     & -19*                    & -23*                    & -                       & -28*                     & -14                     & -20*                    \\
\ourmethod - $\vu_{Entropy, t-1}$                           & -26*                    & -28*                     & -8                      & -23*                    & -                       & -21*                     & -17*                    & -21*                    \\
$P(\vr_{t} | \vu_{t-2}, \vu_{t-1})$ - $\vu_{MaxCI, t-1}$    & -17*                    & \textbf{11}              & -19*                    & -5                      & -                       & \textbf{10}              & -21*                    & -5                      \\
$P(\vr_{t} | \vu_{0:t-1})$ - $\vu_{MaxCI, t-1}$             & -23*                    & -25*                     & -10                     & -21*                    & -                       & -5                       & \textbf{8}              & -18*                    \\
$P(\vr_{t} | \vu_{0:t-1})$ - $\vu_{t-2, t-1}$               & -22*                    & -11                      & -9                      & -13*                    & -                       & -9                       & -16*                    & -15*                    \\
$P(\vr_{t} | \vu_{random}, \vu_{t-1})$ - $\vu_{MaxCI, t-1}$ & -43*                    & -36*                     & -30*                    & -43*                    & -                       & -12*                     & -15*                    & -25*                    \\
\ourmethod - Beam                                           & \textbf{3}              & -2                       & \textbf{1}              & \textbf{3}              & -                       & 5                        & -5                      & \textbf{6}              \\
$P(\vr_{t} | \vu_{t-1})$ - Beam                             & -20*                    & -35*                     & -33*                    & -6                      & -                       & -39*                     & -30*                    & -9                      \\
$P(\vr_{t} | \vu_{0:t-1})$ - Beam                           & -40*                    & -28*                     & -23*                    & -37*                    & -                       & -52*                     & -49*                    & -14*                    \\ \hline
\end{tabular}
}
\caption{The comparisons between inference methods. All models are fine-tuned on \blenderbot.  * indicates a significant difference with our model. "Beam" indicates regularized beam search that employs a width of 10 with 3-grams blocking and a minimum length of 20.}
\label{tab:response generation ablation study}
\end{table*}

\begin{table}[h]
\centering
\resizebox{\linewidth}{!}{
\begin{tabular}{llll}
\hline
\multicolumn{1}{l|}{Models}                     & \textbf{Self-BLEU} $\downarrow$ & \textbf{D-1} $\uparrow$ & \textbf{D-2} $\uparrow$ \\ \hline
\multicolumn{4}{c}{\esconv}                                                            \\ \hline
\multicolumn{1}{l|}{\ourmethod}           & \textbf{0.42}      & \textbf{0.27}       & \textbf{0.74}       \\
\multicolumn{1}{l|}{$P(\vr_{t} | \vu_{t-2}, \vu_{t-1})$}    & 0.69      & 0.27       & 0.70       \\
\multicolumn{1}{l|}{$P(\vr_{t} | \vu_{0: t-1})$} & 0.71      & 0.24       & 0.62       \\
\multicolumn{1}{l|}{$P(\vr_{t} | \vu_{random}, \vu_{t-1})$}        & 0.91      & 0.19       & 0.59       \\ \hline
\multicolumn{4}{c}{\msc}                                                               \\ \hline
\multicolumn{1}{l|}{\ourmethod}           & \textbf{0.69}      & \textbf{0.32}       & \textbf{0.78}       \\
\multicolumn{1}{l|}{$P(\vr_{t} | \vu_{t-2}, \vu_{t-1})$}    & 0.78      & 0.30       & 0.75       \\
\multicolumn{1}{l|}{$P(\vr_{t} | \vu_{0: t-1})$} & 0.80      & 0.27       & 0.74       \\
\multicolumn{1}{l|}{$P(\vr_{t} | \vu_{random}, \vu_{t-1})$}        & 0.93      & 0.20       & 0.53       \\ \hline
\end{tabular}
}
\caption{Response candidates diversity. All models are fine-tuned on \blenderbot}
\label{tab:response_candidates_diversity}
\end{table}

To demonstrate that our method is model-agnostic, we apply our method to \dialogGPT \footnote{\url{https://huggingface.co/microsoft/DialoGPT-medium}} instead of \blenderbot, and evaluate the models on both \esconv and \msc with varying input settings. As you can see from Table~\ref{tab:DialoGPT human evaluation result}, our method outperforms the other \dialogGPT models with different input settings in terms of all metrics. As \dialogGPT uses only a transformer-based decoder, we show that our training and inference methods improve the performance of both decoder-only and encoder-decoder neural architectures.

\paragraph{Ablation Study of Response Generation.}
We conduct ablation studies to demonstrate \textit{conditional dependence} is crucial for selecting direct causes during training and inference. The corresponding results are summarized in Table~\ref{tab:response generation ablation study}.

\begin{table*}[h]
\centering
\resizebox{\textwidth}{!}{
\begin{tabular}{llll}
\hline
\multicolumn{1}{l|}{Human}            & \multicolumn{3}{l}{How long are you doing the online school? (-2, 0, 0, 1)}                                                                                                                                  \\ \hline
\multicolumn{1}{l|}{Input}            & Generated Response                                                                                             & \begin{tabular}[c]{@{}l@{}}Conditional\\ dependence\end{tabular} & Dependence \\ \hline
\multicolumn{4}{c}{\ourmethod (Ours) }                                                                                                                                                                                    \\ \hline
\multicolumn{1}{l|}{$(\vu_0, \vu_5)$} & I am sorry to hear that. What is causing you to be so stressed right now?                                      & 0.0004                                                           & 0.8314     \\
\multicolumn{1}{l|}{$(\vu_1, \vu_5)$} & Do you have any friends or family that you can talk to about it?                                               & 0.0003                                                           & 0.0115     \\
\multicolumn{1}{l|}{$(\vu_2, \vu_5)$} & I am doing well. I am sorry to hear that school is stressful for you right now.                                & 0.0004                                                           & 0.9996     \\
\multicolumn{1}{l|}{$(\vu_3, \vu_5)$} & \begin{tabular}[c]{@{}l@{}}\textbf{I can understand how stressful this is for you. Have you thought about talking to a school counselor?}\end{tabular} & 0.7287                                                           & 0.9287     \\
\multicolumn{1}{l|}{$(\vu_4, \vu_5)$} & Yes, I can understand that. Do you have any friends or family that you can talk to about it?                   & 0.0112                                                           & 0.9981     \\ \hline
\multicolumn{4}{c}{$P(\vr_{t} | \vu_{0:t-1})$ - $\vu_{MaxCI, t-1}$}                                                                                                                                                                    \\ \hline
\multicolumn{1}{l|}{$(\vu_0, \vu_5)$} & I am sorry to hear that. I can understand how stressful it is to be in school right now.                       & 0.0003                                                           & 0.0028     \\
\multicolumn{1}{l|}{$(\vu_1, \vu_5)$} & I’m sorry to hear that. What is stressing you out the most about school right now?                             & 0.0003                                                           & 0.999      \\
\multicolumn{1}{l|}{$(\vu_2, \vu_5)$} & I can understand how stressful school can be. What is stressing you out the most about it?                     & 0.0003                                                           & 0.9472     \\
\multicolumn{1}{l|}{$(\vu_3, \vu_5)$} & \textbf{I can understand how stressful online learning can be, especially online learning.}  (1, 0, -1, -1)                  & 0.872                                                            & 0.7682     \\
\multicolumn{1}{l|}{$(\vu_4, \vu_5)$} & I can understand that. It seems like everyone is stressed about the vaccine.                                   & 0.2121                                                           & 0.9996     \\ \hline
\multicolumn{4}{c}{Other baselines}                                                                                                                                                                                    \\ \hline
\multicolumn{1}{l|}{$P(\vr_{t} | \vu_{t-1})$}            & \multicolumn{3}{l}{It sounds like you're stressed about the way things are going in school. Is that correct? (-3, 0, -3, 1) }   \\
\multicolumn{1}{l|}{$P(\vr_{t} | \vu_{t-2:t-1})$}            & \multicolumn{3}{l}{Have you tried to talk to someone about it?   (-2, 0, -3, -2)}   \\
\multicolumn{1}{l|}{$P(\vr_{t} | \vu_{t-3:t-1})$}            & \multicolumn{3}{l}{I can understand how stressful this is for you. Do you have anyone you can talk to about this?  (-1, 0, 1, -1)}   \\
\multicolumn{1}{l|}{$P(\vr_{t} | \vu_{0:t-1})$}            & \multicolumn{3}{l}{I can understand how stressful it is to be in school right now.  (-2, 1, -3, 0)}   \\
\multicolumn{1}{l|}{$P(\vr_{t} | \vu_{j}, \vu_{t-1})$}            & \multicolumn{3}{l}{I am sorry to hear that. What are you studying?  (-2, -1, -2, -2)}   \\
\multicolumn{1}{l|}{DialoFlow}            & \multicolumn{3}{l}{I'm sorry to hear that. I hope you feel better soon.  (-1, 0, -2, -2)}   \\
\multicolumn{1}{l|}{\begin{tabular}[c]{@{}l@{}}\citep{liu2021emotional} \\ w/o strategy\end{tabular}} & \multicolumn{3}{l}{I'm sorry to hear that. What is stressing you out the most about school right now?  (-1, 0, -1, -2)} \\
\multicolumn{1}{l|}{\begin{tabular}[c]{@{}l@{}}\citep{liu2021emotional} \\ with strategy\end{tabular}} &  \multicolumn{3}{l}{I can understand how stressful school can be. What is stressing you out the most about it?  (-1, 0, -2, 1)} \\
\multicolumn{1}{l|}{Retrieval-guided} &  \multicolumn{3}{l}{I can understand how stressful this is for you. I am sure you are doing your best.  (0, 1, -1,-2)} \\ 
 \hline
\end{tabular}
}
\caption{Response candidates generated by \ourmethod and one baseline $P(\vr_{t} | \vu_{0:t-1})$ based on the conversation history in Table~\ref{tab:spurious_corr_example}. We use $\vu_{MaxCI, t-1}$ to select final responses, which are in \textbf{bold}. Behind responses generated by baselines, we append pair-wise comparison results annotated by five workers between baselines and our model, (Empathy, fluency, informativeness, relevance). In a pair-wise comparison, if baseline is better, it gets a +1 score; if baseline is worse, it gets a -1 score; if baseline is the same with our model, both get 0 score. The sum of the five workers' evaluations is the score shown in this Table.}
\label{tab:generation_example}
\end{table*}

\begin{table*}[h]
\resizebox{\linewidth}{!}{
\begin{tabular}{l|ll}
\hline
\multirow{13}{*}{History} & Supporter & Hi! How are you doing today?  \\ \cline{3-3} 
                         & Seeker    & \begin{tabular}[c]{@{}l@{}}\hl{I am struggling with how to turn in this situation}.\\ \hl{My son is unreasonable but I am trying to help him get through school.} \\ \hl{My boyfriend finds the situation intolerable.}\end{tabular}                                                                                                                       \\ \cline{3-3} 
                         & Supporter & That sounds really hard. Does your son fight with you a lot?                                                                                                                                                                                                                                                                              \\ \cline{3-3} 
                         & Seeker    & Yes, since he got back from NYC he is disrespectful. That is what is upsetting my boyfriend so much.                                                                                                                                                                                                                                      \\ \cline{3-3} 
                         & Supporter & What was he in NYC for? Have you tried getting counseling together?                                                                                                                                                                                                                                                                       \\ \cline{3-3} 
                         & Seeker    & \textbf{\begin{tabular}[c]{@{}l@{}}He went to school there and left school to get acting jobs. He did get his sag and equity cards. \\ He was also working as a caterer. Everything dried up due to Covid. \\ In June, I convinced him to get into therapy but it is slow and I see jealousy with directed at my boyfriend.\end{tabular}} \\ \cline{3-3} 
                         & Supporter & \begin{tabular}[c]{@{}l@{}}Has he gone to therapy with you? Maybe you can try doing some fun activties together.\\  I know my brother had a lot of tension with my mom and finding something they could bond over really helped.\end{tabular}                                                                                             \\ \cline{3-3} 
                         & Seeker    & \textbf{\begin{tabular}[c]{@{}l@{}}\hl{I have tried biking and hiking as well as other activities.  He is openly hostile.} \\ \hl{He does not want anything to do with my boyfriend. He does not want me to go to therapy with him.}\end{tabular}}                                                                                                  \\ \hline
$P(\vr_{t} | \vu_{t-1})$                & Supporter & I can understand why he is hostile to you. Have you tried talking to him about it? (-2, 0, -2, 1)   \\
$P(\vr_{t} | \vu_{t-2:t-1})$            & Supporter & I am so sorry to hear that. I have been in a similar situation and I know how hard it is. (0, 0, -3, -2)  \\
$P(\vr_{t} | \vu_{t-3:t-1})$            & Supporter & Have you tried talking to him about how you feel? Sometimes it helps to talk things through. (-2, 0, -1, 0)  \\
$P(\vr_{t} | \vu_{0:t-1})$              & Supporter & I am so sorry to hear that. Is there anything else I can help you with today? (-2, 0, -2, -3)  \\
$P(\vr_{t} | \vu_{j}, \vu_{t-1})$       & Supporter & I can understand why that would upset you. Have you talked to him about how you feel? (1, 0, -2, 0)  \\
DialoFlow                               & Supporter & I understand how you feel.. It's hard to get away from it. (-1, -1, -2, -1)  \\
\multicolumn{1}{l|}{\begin{tabular}[c]{@{}l@{}}\citep{liu2021emotional} \\ w/o strategy\end{tabular}}   & Supporter & I can understand why that would upset you. Have you talked to him about how you feel? (1, 0, -1, -1)  \\
\multicolumn{1}{l|}{\begin{tabular}[c]{@{}l@{}}\citep{liu2021emotional} \\ with strategy\end{tabular}}  & Supporter & I am so sorry to hear that. I have been in a similar situation before and I know how hard that can be.  (0, 0, -2, -2) \\
Retrieval-guided                        & Supporter & I can understand how stressful this must be for you. Is there anyone you can talk to about this?  (-1, 0, -2, -1) \\
\hline
\ourmethod (Ours)   & Supporter & It sounds like you are trying your best to help your son and your boyfriend at the same time. \\
Human                                   & Supporter & Does he give a reason why?   (-2, 0, -1, 2)    \\
\hline
\end{tabular}
}
\caption{An example where causes of the human response and the generated response partially overlap. The causes of human response are in bold. The causes of the response generated by our model are highlighted. Behind responses generated by baselines, we append pair-wise comparison results annotated by five workers between baselines and our model, (Empathy, fluency, informativeness, relevance). 
}
\label{tab:case_example_part_overlap_causes}
\end{table*}

Training generative models with the utterances selected by our method improves model performance significantly. Without our method, empathy, informativeness and relevance drop for all \blenderbot variations on \esconv. Only the fluency increases slightly when using the preceding two utterances as input during training. It is worth noting that training models with the utterances selected by our CI classifier improves the diversity of response candidates consistently. From Table~\ref{tab:response_candidates_diversity} we can see the diversity of response candidates produced by different response models. The model trained with our method generates more diverse response candidates than the other ones in terms of all metrics. We conjecture that training with direct causes can let model parameters focus on associating key differences among inputs with responses, thus becoming more sensitive to input variations.

Using \blenderbot trained with our method (\ourmethod), we compare our inference method, coined $\vu_{MaxCI, t-1}$, with alternative methods: i) randomly selecting $\vu_{j}$ between $0$ and $t-2$ and combine it with $\vu_{t-1}$, coined $\vu_{Random, t-1}$; ii) taking both $\vu_{t-2}$ and $\vu_{t-1}$ as input, coined $\vu_{t-2, t-1}$; iii) applying the entropy-based method proposed in~\citet{csaky-etal-2019-improving} to remove generic response candidates and select optimal response, coined $\vu_{Entropy, t-1}$; iv) replacing the CI classifier with a dependence classifier for inference, coined $\vu_{MaxDep, t-1}$. The dependence classifier is trained by setting $(\vu_{t-1}, \vr_{t})$ as positive samples, $(\vu_{j}, \vr_{t})$ as negative samples, where $\vu_{j}$ far from responses is randomly sampled from dialogue histories. During inference, we generate response candidates in the same way as our method $\vu_{MaxCI, t-1}$,  but select the candidate that has the highest dependence probability $P_{depend}(l=1|\vu_{j}, \vr_{t}^{j})$ as the final output. 

 The results in Table~\ref{tab:response generation ablation study} show that our inference method outperforms alternative inference methods, when the models are trained with our method. Replacing the CI classifier with the dependence classifier ($\vu_{MaxDep, t-1}$) leads to substantial performance drops in terms of all metrics. It is also noteworthy that generating responses using the preceding two utterances ($\vu_{t-2, t-1}$) is a fairly effective heuristic, which only falls short of our method in terms of empathy. This can be explained by the statistics that $40\%$ of direct causes on \esconv are the preceding two utterances, while the corresponding percentage on \msc is $29\%$. Selecting key utterances randomly or using entropy to pair with $\vu_{t-1}$ is worse than that simple heuristic.
 
 In addition, we compare our method with regularized beam search~\citep{roller2020recipes} in three settings: i) replace the unregularlized beam search with the regularized one using our method, ii) using only preceding utterances as input, and iii) using full conversation histories as input. In all settings, the beam search employs a width of 10 with 3-grams blocking and a minimum length of 20. Regularized beam search with full conversation histories ($P(\vr_{t} | \vu_{0:t-1})$-Beam) or only preceding utterances ($P(\vr_{t} | \vu_{t-1})$-Beam) achieve dramatically lower performance than our inference method. If the beam search is used together with the CI classifier (\ourmethod-Beam), the model performance increases slightly but the differences are not statistically significant. 



\paragraph{Qualitative studies.}  To further investigate the differences between the CI classifier and the dependence classifier, we apply the model to generate all candidate responses and score the candidates with the probabilities yielded by the dependence and the CI classifiers. Using the example conversation in Table~\ref{tab:spurious_corr_example}, we show all generated candidate responses and the corresponding scores in Table \ref{tab:generation_example}. With $\vu_{3}$, the direct cause used by humans, the corresponding response achieves the highest conditional dependence probability but not the highest dependence probability. Perplexity is also not reliable. Moreover, the distributions of the conditional dependence scores are more skewed towards the true direct causes than those of dependence scores. Hence, the conditional dependence, which measures the conditional mutual information obtained from a selected utterance beyond that from the preceding utterance, is more informative and robust than mutual information between responses and single utterances in contexts.

Furthermore, we apply our method to \blenderbot on example dialogues and show qualitative differences to the baselines. Table~\ref{tab:generation_example} shows the responses generated by our method and the baselines using the running example in Table \ref{tab:spurious_corr_example}. The responses generated by our method give a specific suggestion to "talk to a school counselor" or refer to the most specific detail of ``online learning'', while the remaining ones talk about school or irrelevant contents. In addition, we provide the Best-Worse Scaling scores of five crowd-workers, who compare the baseline outputs with those of our method. Most crowd-workers consider our model output is better than that of the baselines in terms of informativeness and relevance. 

For error analysis, we find out that our model cannot always generate natural and relevant responses by relying on the same direct causes as humans. As shown in Table \ref{tab:case_example_part_overlap_causes}, although there are overlapped direct causes between humans and our model, the response generated by our model is reasonable and relevant by capturing context specific entities ``son'' and ``boyfriend'', while the other models fail to do so. In those cases, even if our model uses different direct causes than humans for response generation, most of them are reasonable and fluent. To further investigate to what degree our model utilizes the same direct causes as humans, we apply our model to the test set of \ourdataset and collect the direct causes used during inference. The percentage of using exactly same causes, partially overlapped causes and totally different causes amount to $26.47\%$, $62.13\%$ and $11.40\%$, respectively. Overall, comparing with the baselines, the model with our method produces more specific, relevant, and natural responses than the baselines regardless if it uses the same direct causes as humans or not.

\paragraph{CI Classification Results.}

We evaluate our method \ourmethod to identify direct causes of responses in the test sets of \ourdataset, and compare them with two simple but strong baselines: "Always $\vu_{t-1}$" and "Always $\vu_{t-2}, \vu_{t-1}$". The former always considers $\vu_{t-1}$ of responses as direct causes, while the latter considers the preceding two utterances as direct causes. In the test sets, we keep the manually annotated cause-response pairs as positive examples, while combining all non-cause utterances with $\vu_{t-1}$ and $\vr_{t}$ as negative samples. As a result, the number of negative samples is much larger than the number of positive examples. Due to such an imbalance, we adopt precision, recall, and F1 as the evaluation metrics.

Table~\ref{tab:PC_causal_discovery} reports the results of cause identification. \ourmethod reaches the highest recall and F1 on this task. "Always $\vu_{t-1}$" reaches the highest precision because preceding utterances have the highest probability to be direct causes, as we discussed in Sec.~\ref{sec:annotation of causal graphs}. We also created a balanced test set by randomly sampling non-cause utterances and combining them with $\vu_{t-1}$ and $\vr_{t}$ as negative examples. The accuracy of \ourmethod is $0.83$ on \ourdataset - \esconv, and $0.86$ on \ourdataset - \msc, much higher than random guess.

Furthermore, we evaluate the effectiveness of incremental self-training with constraints on the test sets of \ourdataset by comparing it with three options: i) training only the initial classifier on the labeled training set $\mathbb{D}_L$ of \ourdataset (INIT), ii) fine-tuning the initial classifier on the full unlabeled training set with the context constraint (FC), and iii) incremental self-training without the context constraint on the full unlabeled training set (IST). As shown in Table~\ref{tab:PC_causal_discovery}, \ourmethod outperforms the three options in terms of recall by a wide margin, hence achieves the highest F1 scores on both datasets. Applying the context constraint during self-training filters out mislabeled data far from responses, dropping it leads to the largest reduction of recall and F1. The threshold constraint is still effective by boosting both the precision and the recall of direct cause identification. 


\begin{table}[h]
\centering
\resizebox{\linewidth}{!}{
\begin{tabular}{lccc}
\hline
\multicolumn{1}{l|}{\textbf{Models}}                    & \multicolumn{1}{l}{\textbf{Precision}} & \multicolumn{1}{l}{\textbf{Recall}} & \multicolumn{1}{l}{\textbf{F1}} \\ \hline
\multicolumn{4}{c}{\ourdataset - \esconv}                        \\ \hline
\multicolumn{1}{l|}{Always $\vu_{t-1}$}                   & \textbf{0.80} & 0.41 & 0.54 \\
\multicolumn{1}{l|}{Always $\vu_{t-2}, \vu_{t-1}$}       & 0.60 & 0.61 & 0.61 \\
\multicolumn{1}{l|}{INIT}                   & 0.63 & 0.41 & 0.49 \\
\multicolumn{1}{l|}{FC}        & 0.43 & 0.54 & 0.47 \\
\multicolumn{1}{l|}{IST} & 0.67 & 0.33 & 0.44 \\
\multicolumn{1}{l|}{\ourmethod} & 0.70                         & \textbf{0.71}                       & \textbf{0.70}                   \\ \hline
\multicolumn{4}{c}{\ourdataset - \msc}                           \\ \hline
\multicolumn{1}{l|}{Always $\vu_{t-1}$}                   & \textbf{0.98} & 0.51 & 0.67 \\
\multicolumn{1}{l|}{Always $\vu_{t-2}, \vu_{t-1}$}       & 0.64 & 0.66 & 0.65 \\
\multicolumn{1}{l|}{INIT}                   & 0.70 & 0.60 & 0.65 \\
\multicolumn{1}{l|}{FC}        & 0.49 & 0.59 & 0.54 \\
\multicolumn{1}{l|}{IST} & 0.73 & 0.54 & 0.62 \\
\multicolumn{1}{l|}{\ourmethod}  & 0.73 & \textbf{0.72} & \textbf{0.73} \\ \hline
\end{tabular}
}
\caption{The results of direct cause identification on the test sets of \ourdataset. 
}
\label{tab:PC_causal_discovery}
\end{table}

\section{Related Work}
\paragraph{Dialogue Datasets}
Recently state-of-the-art open-domain dialogue agents have utilized DailyDialog \cite{li-etal-2017-dailydialog}, PersonaChat \cite{zhang2018personalizing}, EmpatheticDialogues \cite{rashkin2019empathetic} and Wizard of Wikipedia \cite{dinan2019wizard}. 
Dialogues in these datasets usually have 3-15 turns. Dialogue agents trained on these dataset don't have the ability to deal with dialogue with very long history.
This weakness encourages researchers to crowdsource long conversations, such as Emotion Support Conversation \cite{liu2021emotional} and Multi-Session Chat \cite {xu2021goldfish}. The number of utterances per dialogue in two datasets is 30 and 53, respectively.

\paragraph{Dialogue Models}
Recently, seq2seq dialogue models, such as DialoGPT, Blenderbot and PLATO~\cite{Zhang2020DialoGPTLG, roller2020recipes, bao2020plato}, showed significant improvement in generating fluent and relevant responses in various dialogue datasets. 
\citet{xu2021goldfish, lewis2021retrievalaugmented, izacard2021leveraging, qu2020rocketqa} propose retrieval-based dialog systems that select relevant utterances from history as input. However, such methods select utterances based on semantic relevance, which may still suffer from spurious correlation in input. \citet{whang2021response, niu2018adversarial, Nyoungwoo9736525, akama2020filtering} seek to first generate or retrieve response candidates, then select final responses using dialog–response binary classifier. Such binary classifiers are trained to identify relevance or irrelevance. However, relevance includes causation and spurious correlation, which cannot be identified by those classifiers.

\section{Conclusion}
We conduct the \textit{first} study from a causal view to investigate and tackle spurious correlations in dialogues. Inspired by constraint-based causal discovery algorithms, we propose a novel constrained self-training method to build a CI classifier by using a small corpus \ourdataset, which is manually annotated with causal graphs by us. The CI classifier is applied to filter out spuriously correlated utterances in conversation histories before training a response generation model. That classifier also serves as a scoring function during inference to select the best response from all generated candidates. By identifying conditionally dependencies between utterances and responses, our model agnostic approach significantly improves the overall generation quality of response models in terms of relevance, informativeness and fluency.

\section*{Acknowledgments}
We thank the Action Editor and the anonymous reviewers for their constructive feedback. This material is based on research sponsored by Air Force Research Laboratory and DARPA under agreement numbers FA8750-19-2-0501 and HR001122C0029. The U.S. Government is authorized to reproduce and distribute reprints for Governmental purposes notwithstanding any copyright notation thereon. The computational resources of this work are supported by the Multi-modal Australian ScienceS Imaging and Visualisation Environment (MASSIVE).

\bibliography{tacl2021}
\bibliographystyle{acl_natbib}



\end{document}